\documentclass[letterpaper]{article} % DO NOT CHANGE THIS
\usepackage[]{aaai2026}  % DO NOT CHANGE THIS
\usepackage{times}  % DO NOT CHANGE THIS
\usepackage{helvet}  % DO NOT CHANGE THIS
\usepackage{courier}  % DO NOT CHANGE THIS
\usepackage[hyphens]{url}  % DO NOT CHANGE THIS
\usepackage{graphicx} % DO NOT CHANGE THIS
\usepackage{booktabs}
\usepackage{natbib}  % DO NOT CHANGE THIS AND DO NOT ADD ANY OPTIONS TO IT
\usepackage{caption} % DO NOT CHANGE THIS AND DO NOT ADD ANY OPTIONS TO IT
\usepackage{algorithm}
\usepackage{amsmath}
\usepackage{algorithmic}
\usepackage{newfloat}
\usepackage{listings}

\usepackage{xcolor}

\usepackage{pgfplots}
\pgfplotsset{compat=1.18}
\usepgfplotslibrary{fillbetween}

\usepackage{newfloat}
\usepackage{listings}
\DeclareCaptionStyle{ruled}{labelfont=normalfont,labelsep=colon,strut=off} % DO NOT CHANGE THIS
\floatstyle{ruled}
\newfloat{listing}{tb}{lst}{}
\floatname{listing}{Listing}
\title{Different Theories, One System:\\The Empirical Organization of Hostile Rhetoric}
\title{Many Theories, One System:\\Unifying Models of Hostile Rhetoric in Online Discourse}
\title{%Many Theories, One System:\\
Testing Theories of Hostile Rhetoric in Online Discourse}
\title{Unifying Models of Intergroup Hostility in Online Discourse}

\author{
    Patrick Gerard\textsuperscript{\rm 1}, 
    Julia Mendelsohn\textsuperscript{\rm 2}, 
    Kristina Lerman\textsuperscript{\rm 3}
}
\affiliations{
    \textsuperscript{\rm 1}Information Sciences Institute, University of Southern California\\
    \textsuperscript{\rm 2}University of Maryland\\
    \textsuperscript{\rm 3}Indiana University   , Bloomington\\
    pgerard@isi.edu, juliame@umd.edu, krlerman@iu.edu
}

\usepackage{bibentry}
\begin{document}

\maketitle

% \begin{abstract}
% Harmful rhetoric is typically studied through distinct mechanisms such as boundary construction, threat, scapegoating, negative evaluation, dehumanization, and calls for harmful action. These mechanisms come from different theoretical traditions and are often examined separately, leaving much unknown about how they actually relate to one another in discourse in both structure and time. Using over 2.86 million posts from TikTok, Truth Social, and Twitter/X during the 2024 U.S. election, we bring six mechanisms drawn from major theoretical traditions into a common empirical framework and compare their organization with the relationships and developmental sequences proposed by existing theories. We find substantial differences in how well existing theories map onto the structural and temporal organization observed in actual discourse. Yet when these mechanisms---which are typically studied in isolation---are brought into a common framework, they form a remarkably regular and robust system, with stable relationships among mechanisms and consistent patterns in when those mechanisms emerge within narratives. These findings suggest that influential theories of harmful rhetoric are best understood as partial accounts of a broader, regularly organized system, with implications for how harmful discourse is theorized, modeled computationally, and identified for intervention.

% \end{abstract}

\begin{abstract}
Hostile rhetoric toward social groups can normalize exclusion and justify mistreatment, as well as contribute to rising polarization and political violence. Efforts to moderate hostile rhetoric in online speech draw on foundational theories in social and moral psychology, and political science. However, these theories were developed largely in parallel, often propose different and sometimes conflicting accounts of how hostility develops, and have rarely been tested against each other in real discourse. The result is a fragmented understanding of the rhetorical mechanisms of hostility,  without a clear sense of how they appear, and relate to each other, in real-world discourse. Using 2.86 million posts from TikTok, Truth Social, and Twitter/X during the 2024 U.S. presidential election, we model the mechanisms of six foundational theories of intergroup hostility -- boundary construction, threat construction, scapegoating, negative evaluation, dehumanization, and action orientation -- within a common empirical framework to recover the broader organization of intergroup hostility rhetoric. Structurally, we find that boundary construction and threat construction anchor the system; temporally, we find that these mechanisms tend to follow a regular ordering: boundary construction, derogation, and action orientation tend to appear early; dehumanization and threat construction later; scapegoating latest. Mapping how these theoretical frameworks actually manifest in discourse bridges longstanding divisions across social science traditions and presents computational social science with a clearer empirical foundation for modeling intergroup hostility rhetoric beyond single-label detection.

\end{abstract}

\begin{links}
    \link{Code}{https://tinyurl.com/harmful-speech-icwsm}
    \link{Data}{https://tinyurl.com/harmful-speech-icwsm}
\end{links}

\section{Introduction}

Online platforms have spent more than a decade trying to identify and moderate harmful language at scale, focusing on slurs, direct threats, and overt calls for violence. Much of the discourse that targets social groups, however, is less explicit: speakers may draw a boundary between ``us'' and ``them,'' portray an outgroup as dangerous, blame it for social problems, describe its members in degrading or dehumanizing terms, or argue that exclusionary or punitive treatment is warranted~\citep{reicher2008making,cap2008towards,cap2013proximization,glick2002sacrificial,haslam2006dehumanization,benesch2012dangerous}. Such rhetoric may avoid overtly hateful or toxic language while still contributing to negative outcomes such as affective polarization,  prejudice, and targeted violence.

Computational research has often grouped hostile rhetoric into broad categories such as hate speech, toxicity, or abuse, even though those categories are defined inconsistently across datasets and studies and often capture substantively different phenomena~\citep{fortuna2018survey,schmidt2017survey,davidson2017automated,fortuna2020toxic,vidgen2020directions}.
Part of this heterogeneity reflects the distinct psychological and social processes that can be expressed through group-directed hostile rhetoric. Influential theories of intergroup hostility, exclusion, and political threat have developed across social psychology, moral psychology, intergroup relations, and political science~\citep{tajfel1979integrative, turner1987rediscovering, reicher2008making, opotow1990moral, haslam2006dehumanization, glick2002sacrificial, cap2008towards, cap2013proximization}. These traditions have shaped how scholars understand how outgroups are constructed, threats are perceived, blame is assigned, moral standing is eroded, and harmful treatment is legitimized.

These theories have, in turn, strongly influenced computational social science, shaping how researchers define harmful speech and operationalize constructs such as hate speech, dehumanization, othering, threat, and incitement~\cite{elsherief2018hate, salminen2018anatomy, silva2016analyzing, mendelsohn2020framework, gerard2025fear}. Yet different theoretical traditions emphasize different psychological mechanisms, which are usually studied separately. Some accounts foreground group differentiation and threat; others focus on blame, derogation, moral exclusion, dehumanization, or harmful action. They also propose different relationships among these mechanisms and, in some cases, different sequences in which they are expected to develop. We are thus left with a fragmented picture of group-directed hostile rhetoric: a set of well-developed theories and computational measures, but no common empirical account of how their mechanisms fit together in actual discourse, whether their proposed relationships reflect the same underlying organization, or where the theories diverge.

Using 2.86 million pieces of social media content from three platforms during the 2024 U.S. presidential election, we bring six mechanisms from these traditions into a common empirical framework: boundary construction, threat construction, scapegoating, negative evaluation, dehumanization, and action orientation. We evaluate theory-derived structural relationships using held-out predictive dependence and conditional association, and test theory-derived temporal sequences using mechanisms' first appearances within cross-platform narratives. We then recover the broader structural and temporal organization that emerges when all six mechanisms are considered together.

We find a robust empirical system with a core centered on boundary construction, threat construction, and action orientation. The mechanisms also show a regular temporal pattern, with boundary construction, negative evaluation, and action orientation tending to appear early in discourse; dehumanization and threat construction later; and scapegoating latest.

Influential theories of group-directed hostile rhetoric thus appear to describe different parts of the same underlying system. No single account captures the system in full, but their respective mechanisms have a regular structural and temporal organization when examined together. A unified empirical account of that organization connects theoretical traditions that have largely developed in parallel and gives computational work a clearer basis for modeling group-directed hostile rhetoric. For platforms and online communities, this offers a way to track hostile rhetoric earlier in its development and distinguish where different forms of moderation are most appropriate.

\section{Related Work}

Research on group-directed hostile rhetoric spans several major theoretical traditions, each emphasizing different processes through which outgroups become targets of hostility. These traditions identify mechanisms including group differentiation, perceived threat, blame, derogation, moral exclusion, dehumanization, and the legitimation of harmful action. Although  largely developed in parallel, they have been consequential  within their respective disciplines and have increasingly shaped how computational social scientists conceptualize and measure harmful and hostile discourse online. We focus on these influential accounts and the distinct mechanisms they propose.

\vspace{3pt}
\noindent
\textbf{Group Boundaries, Threat, and Action.}
A recurring theme across theoretical frameworks is that hostility is easier to justify once an outgroup has first been made socially distinct and then cast as a source of danger. \textit{Social Identity Theory} provides a foundation for this sequence by explaining how ingroup--outgroup boundaries are constructed and made salient~\citep{tajfel1979integrative,turner1987rediscovering}. \textit{Intergroup Threat Theory} complements this account by identifying perceived realistic and symbolic threats to the ingroup as key drivers of hostility towards outgroups~\citep{stephan2009intergroup}. The \textit{Five-Step Social Identity Model} makes this developmental logic more explicit, linking identification and exclusion to the construction of threat and, ultimately, to the legitimation of measures against the outgroup~\citep{reicher2008making}. \textit{Proximization} theory arrives at a related logic from a different tradition, showing how representations of an external threat can move discourse toward the justification of preventive or coercive action~\citep{cap2008towards,cap2013proximization}. Across these accounts, boundary construction and threat constitute distinct but related processes through which hostility can develop and mistreatment become normalized or legitimized.

\vspace{3pt}
\noindent
\textbf{Moral Exclusion and Dehumanization.}
Other influential traditions focus on the moral and human status accorded to the target group. \textit{Moral exclusion theory} describes the process by which individuals or groups are placed outside the community to which ordinary norms of fairness and moral obligation apply~\citep{opotow1990moral}. \textit{Dehumanization} describes a related process in which targets are denied or diminished in their humanity, agency, individuality~\citep{haslam2006dehumanization,bandura1999moral}. \textit{Infrahumanization} extends this logic to subtler forms of exclusion, in which outgroups are attributed fewer uniquely human qualities or emotions~\citep{haslam2014dehumanization}. Across these accounts, the outgroup is not simply portrayed as dangerous, but its moral and human status is narrowed in ways that can weaken normative constraints against mistreatment.

\vspace{3pt}
\noindent
\textbf{Blame, Derogation, and Harmful Action.}
A third set of traditions focuses on how hostility becomes attached to a target through blame, negative judgment, and the legitimation of harmful treatment. \textit{Scapegoating}  examines how responsibility for social problems, crises, or decline is attributed to particular groups~\citep{glick2002sacrificial}, while frameworks of \textit{negative evaluation} capture derogatory judgments that are distinct from both perceived threat and outright dehumanization~\citep{fiske2018model}. At the more action-oriented end of the spectrum, \textit{dangerous-speech} and \textit{incitement} frameworks examine rhetoric that advocates, legitimates, or encourages harmful treatment of a target~\citep{benesch2012dangerous}. Across these accounts, the emphasis shifts from defining the outgroup or its moral status to assigning it culpability and negative worth so as to justify mistreatment.

\vspace{3pt}
\noindent
\textbf{Computational Approaches to Harmful Speech.}
These theories have shaped how computational social scientists conceptualize hate speech and related forms of rhetoric. Instead of relying on self-reports and survey instruments, early computational efforts focused on identifying relevant constructs in the text of online posts, specifically hate speech, toxicity, and  abuse~\citep{fortuna2018survey,schmidt2017survey,vidgen2020directions}. Subsequent research has separated directed from generalized hate~\citep{elsherief2018hate}, distinguished types and targets of harmful speech~\citep{salminen2018anatomy,silva2016analyzing}, and operationalized more specific constructs such as othering~\citep{gerard2025fear} and fear speech~\cite{saha2023rise}. Other works have examined dynamics of hateful discourse and how it changes around platform interventions or real-world events~\citep{hickey2023auditing,olteanu2018effect}.

% Our work builds directly on these moves toward finer-grained and dynamic analysis, but asks a different question. Rather than studying one harmful-speech category, mechanism, or temporal change at a time, we bring mechanisms drawn from several influential theoretical traditions into the same empirical framework and test how they relate to one another in both structure and time. To our knowledge, prior computational work has not systematically compared these theoretical accounts within the same empirical framework, leaving unclear where they align, where they diverge, and how well they describe observed discourse.
% Despite these advances, important gaps remain. Theoretical mechanisms such as othering, threat, blame, dehumanization, moral exclusion, and calls for harmful action are typically studied in isolation, leaving unclear whether they constitute empirically distinct forms of rhetoric, how they relate to one another, and whether these relationships and mechanisms are visible in naturally occurring discourse. Computational studies likewise tend to focus on individual harmful-speech constructs rather than comparing multiple theoretically grounded mechanisms within a common measurement framework. We address these gaps by developing a set of NLP instruments that operationalize distinct mechanisms drawn from these theoretical traditions and apply them to analyze political narratives spreading on social media. This allows us to test where the mechanisms overlap or diverge, how they co-occur, and how their prevalence and relationships evolve as narratives unfold over time.

Despite these advances, computational work has largely examined harmful-rhetoric mechanisms in isolation. Studies have developed methods for identifying constructs such as hate speech, dehumanization, othering, and fear speech, but these mechanisms are rarely measured together in the same corpus. As a result, we know much less about whether they remain empirically distinct when observed side by side, which relationships among them are most important, and their temporal ordering as narratives develop. To address this gap, we operationalize six theoretically grounded mechanisms within a common measurement framework and analyze their structural and temporal organization across large-scale social media discourse.

% Despite these advances, computational work has largely examined these forms of harmful rhetoric one at a time. Prior studies have developed methods for identifying hate speech~\citep{davidson2017automated}, dehumanization~\citep{mendelsohn2020framework}, othering~\citep{gerard2025fear}, and fear speech~\citep{saha2023rise}, but these mechanisms have rarely been measured together within the same corpus. As a result, we still know little about their interdependencies or the order in which they emerge as narratives develop. We address this gap by placing six theoretically grounded mechanisms in a common measurement framework and analyzing both their structural relationships and temporal organization in naturally occurring discourse.

\section{Data and Measurement}
\subsection{Dataset}

To examine mechanisms of intergroup hostility in online discourse, we study a large, multi-platform corpus spanning a major political event. The corpus contains 2,861,992 annotated posts from 768,438 platform-specific users on TikTok, Twitter/X, and Truth Social during the 2024 U.S. presidential election~\citep{pinto2025tracking,balasubramanian2024public,shah2024unfiltered} (Table~\ref{tab:dataset}). Mechanism prevalence varies somewhat across platforms, although these differences are small in magnitude (see Appendix for platform-specific prevalence and heterogeneity tests).

Beyond individual posts, we also leverage the narrative groupings produced by a method that organizes semantically equivalent cross-platform discourse around the same underlying claims, events, or stories as they develop over time~\cite{gerard2026cross,gerard2026bridging}. This follows the approach used by prior works on computational narrative tracking, in which a narrative is treated as a collection of posts that focus on the same underlying issue, event, or claim~\citep{hanley2024specious, leban2014event, miranda2018multilingual, gerard2025modeling}. This models narratives as coherent units of discourse that can evolve over time and across platforms. For example, one narrative in our corpus follows the attempted assassination of Donald Trump by Ryan Routh~\footnote{\url{https://www.justice.gov/opa/pr/ryan-wesley-routh-sentenced-life-prison-attempted-assassination-president-donald-j-trump-and}}, and links TikTok posts speculating about a broader conspiracy, Twitter/X posts discussing his attempted-assassination charge, and Truth Social posts identifying and discussing his role as the alleged gunman. 

Our temporal analysis identifies 2,513 narratives, with 99.2\% of these spanning at least two of the three platforms. Prior validation shows that these narratives remain coherent across platform-specific forms of expression and support cross-platform analysis (we discuss and validate this further in the Appendix).

% Beyond individual posts, we also leverage the narrative groupings produced by a cross-platform method \cite{gerard2026cross,gerard2026bridging}, which organizes semantically equivalent cross-platform discourse around the same underlying claims, events, or stories as they develop over time. Prior validation shows that these narratives remain coherent across platform-specific forms of expression and support cross-platform analysis of narratives (see Appendix%~\ref{tk} 
%  for data details).

This allows us to study hostile rhetoric at two levels: the relationships among mechanisms across individual pieces of discourse, and the order in which those mechanisms emerge as narratives unfold.

\begin{figure*}[t]
    \centering
    \includegraphics[width=2.1\columnwidth]{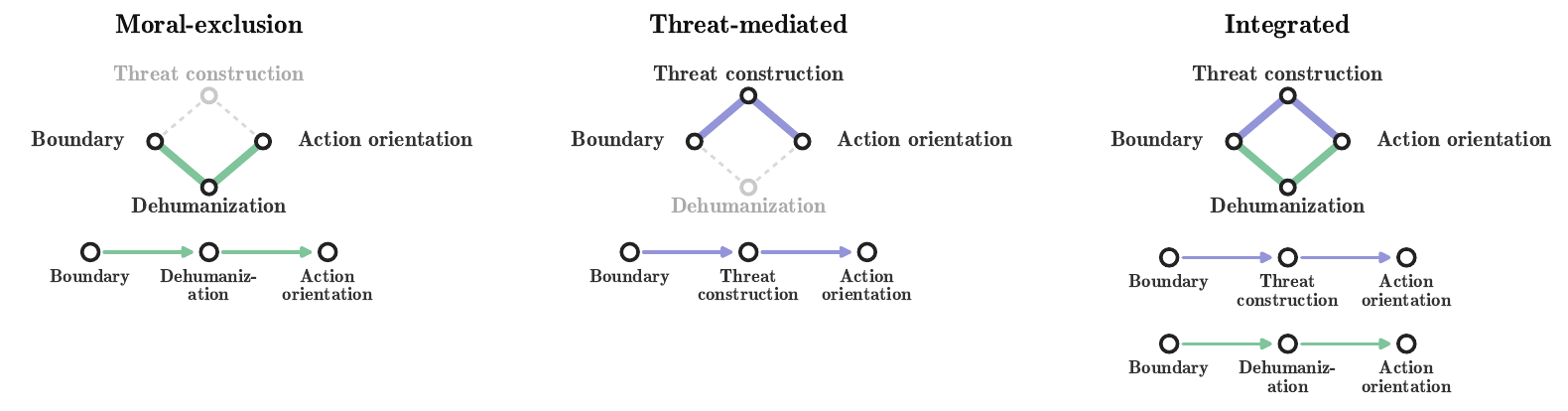}

\caption{
Theory-constrained models used for comparison. \textbf{Diamonds show the structural relationships} each model permits: the moral-exclusion and threat-mediated models encode the relationships implied by their respective theoretical accounts, the integrated model combines both, and the independent model contains none. The integrated model therefore tests whether the two theoretical accounts capture complementary parts of the same empirical organization. Below each structural model, \textbf{chains show the corresponding predicted temporal order} in which the mechanisms should appear.}
\label{fig:theory-schematic}
\end{figure*}

\begin{table}[h]
\centering
\caption{Composition of the social media corpus by platform. Percentages indicate each platform's share of the full corpus.}
\label{tab:dataset}
\begin{tabular}{lr}
\toprule
\textbf{Platform} & \textbf{Posts} \\
\midrule
TikTok       & 807,766 (28.2\%) \\
Twitter/X    & 1,692,711 (59.1\%) \\
Truth Social & 361,515 (12.6\%) \\
\midrule
\textbf{Total} & \textbf{2,861,992} \\
\bottomrule
\end{tabular}
\end{table}

\subsection{Mechanism Definitions}

Throughout this paper, we use \emph{group-directed hostile rhetoric} as an umbrella term for the mechanisms we study, broadly referring to discourse that constructs, evaluates, blames, threatens, degrades, or advocates harmful treatment toward a social target.
%This broader rhetorical space subsumes 
%using the mechanisms studied here. 
We note that our labels capture the linguistic expression of the mechanisms. Threat construction, for example, identifies rhetoric that presents a group as threatening, whether that language reflects the speaker’s own perception, attempts to shape an audience’s perception, or both.

We operationalize six mechanisms based on the theoretical and computational traditions reviewed above. We do not include a separate binary ``hate speech'' label, treating hate speech instead as a broader category that can encompass multiple forms and mechanisms of group-directed hostile rhetoric~\citep{fortuna2018survey,fortuna2020toxic,vidgen2020directions}.
Examples are drawn from political discourse and used for illustrative purposes only. They do not reflect the authors' own views.

\begin{description}
    \item[\textbf{Boundary construction}] Rhetoric that constructs or reinforces a salient ingroup--outgroup distinction, positioning a target as socially distinct from the ingroup~\citep{tajfel1979integrative,reicher2008making}. For example, a post may define a target as not belonging to ``us'' or as fundamentally outside the relevant social group. 
    For example, “Transgender individuals who identify as women are actually men.”

    \item[\textbf{Threat construction}] Rhetoric that represents a target as a threat to the ingroup's security, identity, values, resources, status, or future~\citep{stephan2009intergroup,cap2008towards,cap2013proximization}. A post may portray a target as endangering people's safety, institutions, culture, or way of life.
Example: ``Trans women in sports isn't inclusion — it's the end of women's sports as we know it.''

    \item[\textbf{Scapegoating}] Rhetoric that assigns responsibility or blame to a target for social problems, crises, decline, or other harmful conditions~\citep{glick2002sacrificial}. For example, a post may attribute economic decline, crime, political instability, or other social problems to a particular group. 
    % Example: “While we can barely afford healthcare, government is paying for trans-gender illegal aliens to have sex changes in prison.”
    Example: "We can't afford healthcare for our own citizens while billions of dollars go to Israel."

    \item[\textbf{Dehumanization}] Rhetoric that denies or diminishes a target's humanity, individuality, agency, moral standing, or other uniquely human characteristics~\citep{haslam2006dehumanization,bandura1999moral,haslam2014dehumanization}. For example, a target may be described as animalistic, contaminating, subhuman, or otherwise outside ordinary human or moral consideration.
    Example: "In Springfield, they're eating the dogs. The people that came in, they're eating the cats."

    \item[\textbf{Negative evaluation}] Rhetoric that expresses derogatory, contemptuous, and strongly negative judgments of a target without necessarily portraying it as threatening or less human~\citep{fiske2018model}. For example, a post may characterize a target as disgusting, immoral, stupid, weak, or otherwise contemptible.
 Example: "They're not sending their best---they're sending criminals, drug dealers, and rapists."
    
    \item[\textbf{Action orientation}] Rhetoric that advocates, legitimates, or normalizes harmful action or treatment of target~\citep{benesch2012dangerous,reicher2008making}. For example, a post may call for a target to be excluded, punished, removed, imprisoned, deported, or otherwise subjected to harmful treatment.
    Example: "Criminal illegal aliens should be detained and deported."

\end{description}

\subsection{Annotation and Validation}
\label{sec:annotation_validation}

We annotated the six mechanisms at scale using a human-validated LLM pipeline adapted from prior work on theory-grounded harmful-speech annotation~\citep{gerard2025fear}. Since positive instances are relatively rare, we first draw a large sample stratified by platform and time and use GPT-5 to identify likely positive and negative cases for each mechanism; these labels were used only for sampling and are never shown to the human annotators.

Two trained annotators then independently labeled each sampled post--mechanism pair (100 of each) using the definitions above  (the full annotation guidelines are provided in the repository). Mean mechanism-level Cohen's $\kappa$ before adjudication is .643, comparable to agreement reported for related harmful-language annotation tasks~\citep{matter2024investigating,gerard2025fear}. Disagreements were adjudicated and unresolved cases excluded. Then, to obtain enough positives of each class, additional GPT-5-positive candidates were sampled to obtain sufficient confirmed positives, after which excess negatives were subsampled to produce 100 evaluation items per mechanism.

The final benchmark contains 600 post--mechanism pairs, with 262 positives and 338 negatives, spanning Twitter/X (377; 62.8\%), TikTok (156; 26.0\%), and Truth Social (67; 11.2\%) (Table~\ref{tab:validation_composition}). We note that this benchmark is used only to assess annotation performance and not to estimate corpus prevalence.

We evaluate Gemma-3-1B, Gemma-3-4B, Gemma-3-12B, Llama-3.1-8B-Instruct, and Qwen3-14B against the adjudicated labels using precision, recall, F$_1$, and Cohen's $\kappa$~\citep{gerard2025fear,heseltine2024large,matter2024investigating}. Qwen3-14B performs best overall, achieving macro-F$_1=.798$ and mean mechanism-level $\kappa=.588$ (Table~\ref{tab:llm-performance}), and is therefore used for corpus-scale annotation. Since downstream analyses rely on these automated labels, we also test whether the recovered structural organization is robust to plausible annotation error using prevalence-preserving label perturbations (see Appendix for details).

\begin{table}[ht]
\centering
\small
\caption{Candidate LLM performance against the human-adjudicated reference set. Metrics are macro-averaged across the six mechanisms.}
\label{tab:llm-performance}
\begin{tabular}{lrrrrr}
\toprule
Model & Accuracy & Precision & Recall & F$_1$ & $\kappa$ \\
\midrule
Gemma-3-1B      & .443 & .439 & .976 & .600 & .011 \\
Llama-3.1-8B    & .502 & .466 & .995 & .630 & .102 \\
Gemma-3-4B      & .602 & .519 & .938 & .665 & .247 \\
Gemma-3-12B     & .715 & .609 & \textbf{.944} & .737 & .440 \\
Qwen3-14B       & \textbf{.797} & \textbf{.700} & .931 &
                  \textbf{.798} & \textbf{.588} \\
\bottomrule
\end{tabular}
\end{table}

\subsection{Theory-Derived Expectations}

Existing theories provide concrete expectations about which mechanisms of hostile rhetoric should be related and, in some cases, how they should unfold over time. We use these expectations as reference points for interpreting the empirical organization of the six mechanisms. Because we use language data, we evaluate these expectations through their observable rhetorical manifestations rather than attempting to measure the underlying psychological processes directly. Figure~\ref{fig:theory-schematic} illustrates the proposed structural and temporal organization of each theory.

\begin{description}

    \item[\textbf{Threat-mediated pathway}] 
    Social identity, collective-hate, and proximization theories place group differentiation upstream of threat construction, with threat in turn helping legitimate defensive, preventive, or coercive action. %~\citep{reicher2008making,cap2008towards,cap2013proximization}. 
    This implies both structural relationship (between boundary construction and threat construction, between threat construction and action orientation) and temporal relationship, with  mechanisms emerging in temporal order (boundary construction $\rightarrow$ threat construction $\rightarrow$ action orientation).

    \item[\textbf{Moral-exclusion pathway}] 
       Moral exclusion and dehumanization theories describe a different logic in which targets are progressively removed from ordinary moral concern or denied full human standing, making harmful treatment easier to justify. %~\citep{opotow1990moral,haslam2006dehumanization,bandura1999moral}. 
       We therefore expect boundary construction and negative evaluation to be associated with dehumanization, dehumanization to be associated with action orientation, and dehumanization to emerge downstream of group differentiation and derogation but prior to harmful action.

    \item[\textbf{Integrated pathway}] 
    These accounts need not be mutually exclusive. Threat-based mobilization and moral exclusion may represent distinct but connected routes through hostile rhetoric, with group differentiation feeding both. We therefore also test an integrated structure containing the relationships implied by both accounts and assess whether the two pathways capture complementary aspects of the empirical organization of hostile rhetoric.
\end{description}

\section{Analytical Framework}

Our analysis examines how the mechanisms of group-directed hostile rhetoric are organized in online discourse and asks how well prominent theoretical accounts correspond to that empirical organization. We study this organization in two dimensions: the \emph{structural} relationships among mechanisms within discourse and their \emph{temporal} ordering as narratives evolve.

\subsection{Measuring Structural Organization}

Our structural analysis asks which relationships among the six mechanisms remain informative once the other mechanisms are taken into account. We examine the structure from three complementary perspectives: held-out predictive information, conditional association, and independently learned Bayesian-network structure. These approaches make different assumptions and quantify dependence differently; we treat convergence across these analyses, then, as evidence that the recovered relationships are robust features of the data and not an artifact of a particular statistical representation. Full estimation details are provided in the Appendix.

Our primary structural analysis uses held-out prediction. For each candidate structure, we predict each mechanism from the mechanisms connected to it and evaluate performance on held-out posts using conditional log-likelihood. We then compare each theory-derived structure against an independent structure with no edges and a benchmark containing all 15 pairwise relationships, and report how much of the available predictive gain it recovers. Since predictive fit can improve simply by adding more relationships, we compare each theory-derived structure against every alternative structure with the same number of edges; this allows us to distinguish informative relationships from simply a larger edge budget.

We then recover the empirical structure directly from the data using the same three complementary views of dependence. First, for each pair of mechanisms, we measure how much one improves prediction of the other after accounting for the remaining four mechanisms; averaging this contribution in both directions gives a single edge weight. We rank all 15 relationships by this contribution and define the empirical backbone as the smallest set of edges that captures at least 95\% of the total positive predictive weight. Second, we estimate multivariate logistic models for each mechanism while conditioning on the other five and report the resulting conditional odds ratios. Third, we independently learn Bayesian-network structures from the joint distribution of the six mechanisms using BIC-scored hill-climbing and compare their undirected skeletons with the predictive backbone. These analyses let us ask whether the same relationships emerge under predictive, conditional-association, and graphical-model views of the system.

All structural analyses use five-fold held-out cross-validation, with empirical edge weights estimated within training data using stratified inner folds. We repeat structure recovery across folds, separately by platform, and under prevalence-preserving perturbations of the automated labels (see Appendix for details).

\subsection{Measuring the Temporal Ordering of Mechanisms Within Narratives}

We next examine whether the temporal relationships proposed by existing theories correspond to how mechanisms unfold within narratives. For each theoretical pathway, we evaluate both the full sequence and its constituent pairwise orderings among narratives in which all relevant mechanisms appear. This distinction allows us to separate evidence for individual transitions from evidence for the pathway as a whole. We report the proportion of narratives exhibiting each predicted pairwise ordering and the complete sequence, and compare the observed full-sequence rate with a permutation null that preserves the observed first-appearance positions within each narrative while randomly reassigning those positions among the mechanisms in the pathway. 

We then examined whether a broader temporal pattern structures all six mechanisms simultaneously. As Figure~\ref{fig:temporal-ranking} shows, three distinct tiers emerge across 97.5\% of bootstrap samples: boundary construction, negative evaluation, and action orientation tend to appear early; dehumanization and threat construction follow; and scapegoating consistently appears latest. Notably, mechanisms that are strongly connected in the structural analysis can occupy different positions in narrative development, while mechanisms with similar timing are not necessarily the most tightly linked structurally (note that these patterns describe the typical ordering of first appearances across narratives and should not be read as causal steps). 

Table~\ref{tab:guiding-example-temporal} illustrates this pattern in an anti-trans narrative. The narrative begins by drawing a sharp boundary around who counts as a woman and attaching negative judgments to those outside it. Calls to exclude or restrict transgender people appear alongside this early framing. Threat claims tend to emerge later, recasting the outgroup as dangerous, followed by more explicit dehumanizing language and broader blame directed at institutions and groups portrayed as responsible for the perceived harm. The example reflects the same tiered pattern seen across the corpus: boundary setting, evaluation, and action tend to enter early, threat and dehumanization later, and scapegoating latest.

\section{Results}

\subsection{Existing Theories Differ in Structural Fit}
\begin{figure}[ht]
    \centering
    \includegraphics[width=.95\columnwidth]{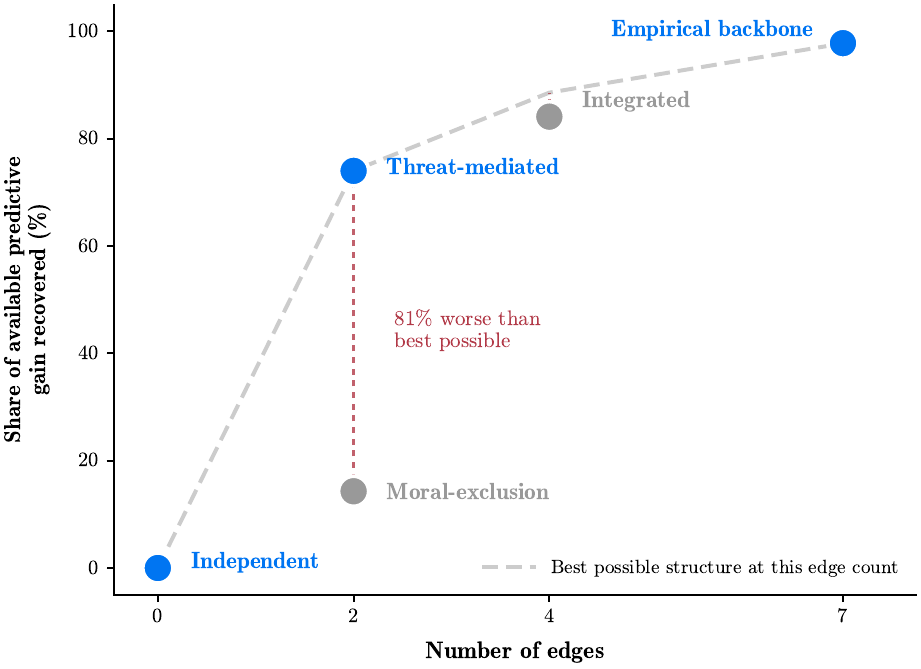}
    \caption{Predictive gain recovered as a function of edge budget. The dashed line traces the best-performing structure at each edge count. The threat-mediated model recovers more predictive information than any other two-edge structure, while the moral-exclusion model performs substantially worse.}
    \label{fig:efficient-frontier}
\end{figure}

The structural relationships proposed by existing theories differ markedly in levels of empirical support across datasets (Fig.~\ref{fig:efficient-frontier}). The threat-mediated pathway performs especially well: with only two edges, it recovers 74.0\% of the predictive gain available to the fully connected benchmark, making it the best-performing of all 105 possible two-edge structures. By contrast, the moral-exclusion pathway recovers only 14.3\% of the available gain, ranking 40th of the 105 two-edge structures (62.9th percentile). %, achieving just 19.3\% of the predictive performance of the best two-edge structure.

Combining the two theories improves overall fit: the integrated four-edge structure recovers 84.1\% of the available predictive gain and ranks seventh among 1,365 possible four-edge structures. However, most of this performance is already captured by the two threat-mediated relationships---boundary construction--threat construction and threat construction--action orientation---which are among the most informative dependencies in the system; the additional moral-exclusion relationships contributes substantially less predictive gain. Thus, the empirical organization of hostile rhetoric aligns most strongly with social-identity and threat-based accounts that place group differentiation and perceived threat at the center of mobilization against an outgroup~\citep{reicher2008making,cap2008towards,stephan2009intergroup,cap2013proximization}. Empirically, these relationships appear to capture a particularly prominent feature of the organization of hostile rhetoric.

\subsection{Local Relationships Form a Stable Global Structure}

\begin{figure}[t]
    \centering
    \includegraphics[width=.88\columnwidth]{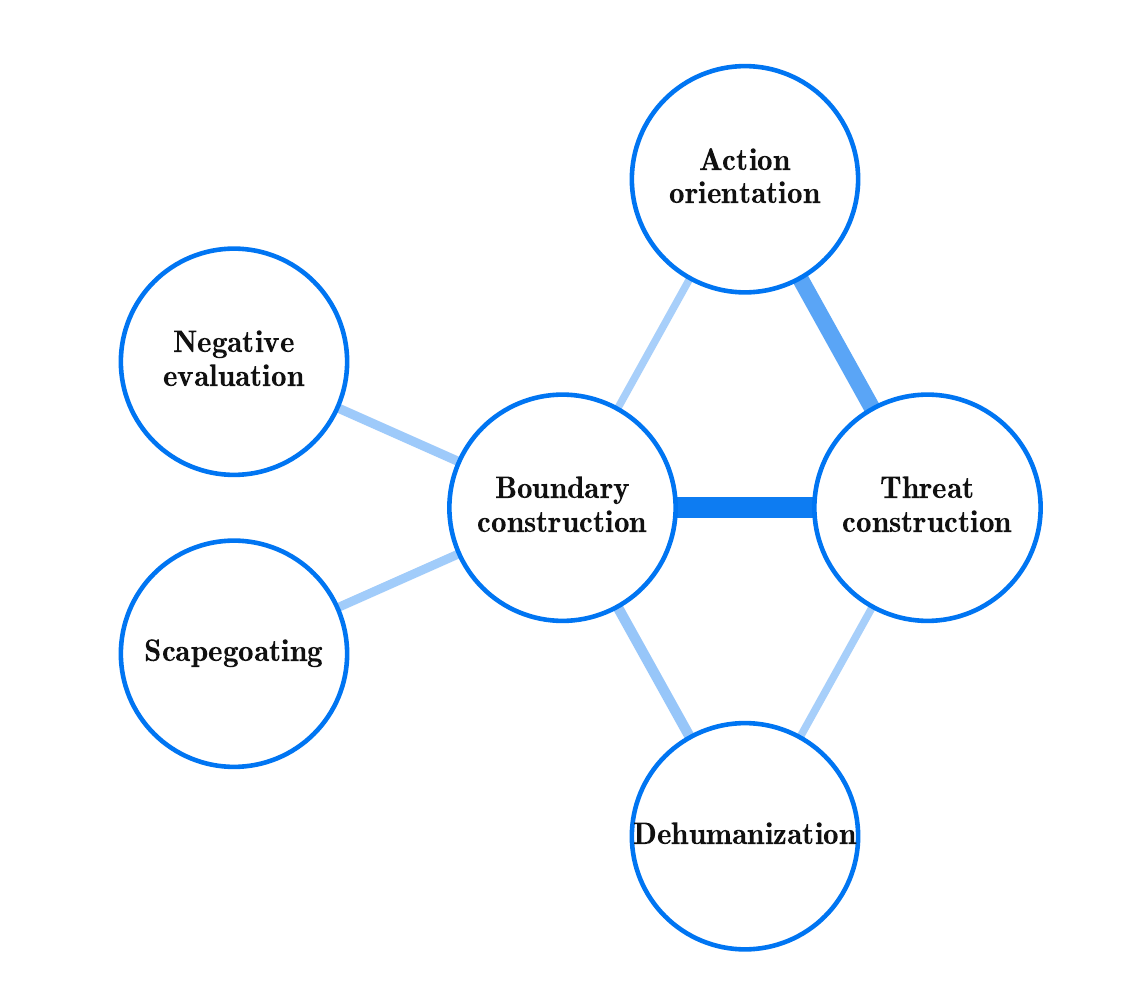}
    \caption{
    Empirical backbone of the mechanisms of group-directed hostile rhetoric. The structure recovers nearly 98\% of the predictive gain available to the fully connected system. Edge width and color intensity both indicate the strength of each relationship, with boundary construction–threat construction the strongest in the system. Edges represent conditional relationships between mechanisms within the same post.
}
    \label{fig:backbone-structure}
\end{figure}

The contrast between theories becomes clearer when we examine the individual relationships that compose the system. As shown in Figure~\ref{fig:backbone-structure}, the two strongest relationships are those specified by the threat-mediated account: the  boundary construction--threat construction relationship accounts for 47.0\% of the potential conditional predictive information, while the  threat construction--action orientation relationship contributes another 25.2\%. The same concentration appears under a distinct measure of dependence: in the conditional logistic models (Appendix Figure~\ref{fig:conditional-odds}), boundary construction and threat construction are associated with approximately a twenty-fold change in one another's odds after conditioning on the remaining mechanisms, while threat construction and action orientation are associated with approximately a fifteen-fold change.

% Beyond these two dominant relationships, the conditional odds show that the system is broader. When boundary construction is present, the odds of scapegoating are about 15 times higher, the odds of dehumanization about 9 times higher, and the odds of negative evaluation about 7 times higher, even after accounting for the other mechanisms. 

These local relationships assemble into a sparse global structure (Figure~\ref{fig:backbone-structure}). Fewer than half of all possible relationships recover 97.8\% of the predictive information available in the fully connected system. Boundary construction and threat construction anchor this backbone: boundary construction connects broadly across mechanisms, while threat construction participates in the two strongest dependencies in the system. The resulting organization is also highly stable. The same seven relationships are recovered across cross-validation folds, all appear in the independently learned Bayesian-network skeleton, and the same core structure emerges when the analysis is repeated separately on TikTok, Truth Social, and Twitter/X. The relative strength of these relationships is also robust to substantial perturbation of the automated labels (see Appendix for details).

\subsection{Temporal %Ordering Follows a Different 
Organization 
of Mechanisms}

We next ask whether the developmental sequences proposed by existing theories are reflected in the temporal ordering of mechanisms within narratives (see second row of Figure~\ref{fig:theory-schematic} for examples). Direct evaluation of the theory-derived pathways provides only partial support. The threat-mediated sequence---boundary construction, followed by threat construction, followed by action orientation---occurs in 14.3\% of eligible narratives, significantly more often than under the permutation null (11.7\%, $p=.002$). However, this support is concentrated in the first step: boundary construction precedes threat construction in 68.9\% of non-tied cases, whereas threat construction precedes action orientation in only 40.1\%.

The moral-exclusion pathway---boundary construction, followed by dehumanization, followed by action orientation---is not recovered as a complete sequence. It appears in 14.5\% of eligible narratives, essentially identical to the permutation expectation (14.5\%, $p=.538$). Here too, the first step is strongly supported: boundary construction precedes dehumanization in 77.0\% of non-tied cases, but dehumanization precedes action orientation in only 29.0\%.

\begin{figure}[t]
    \centering
    \hspace*{-0.095\columnwidth}
    \includegraphics[width=1.12\columnwidth]{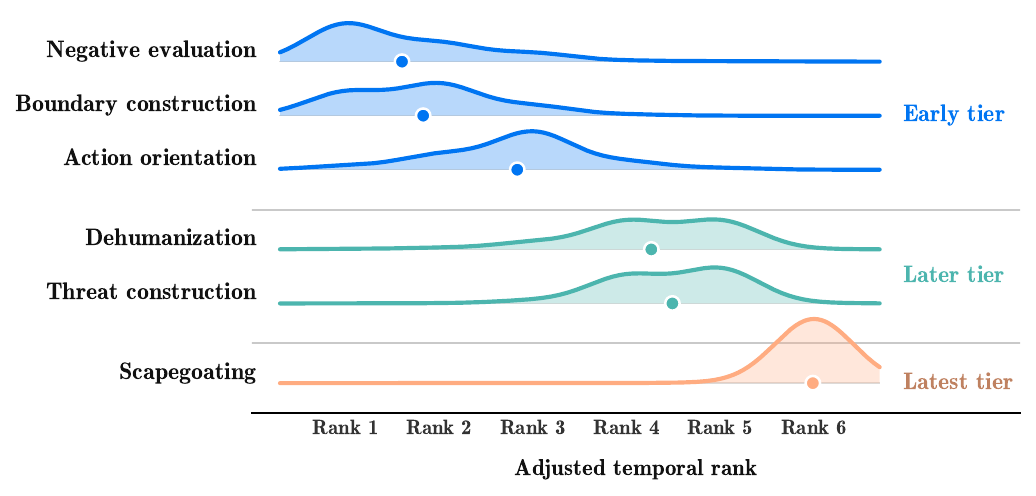}
    \caption{
    Bootstrap distributions of temporal rank after accounting for mechanism frequency and narrative length. Negative evaluation, boundary construction, and action orientation form an early tier; dehumanization and threat construction form an intermediate tier; and scapegoating remains distinctly late.
    }
    \label{fig:temporal-ranking}
\end{figure}

We next ask whether a broader temporal organization emerges when all six mechanisms are considered together. As shown in Figure~\ref{fig:temporal-ranking}, the mechanisms differ systematically in where they tend to first appear in the narratives. Boundary construction, negative evaluation, and action orientation tend to enter narratives earlier; dehumanization and threat construction tend to first appear later; and scapegoating consistently appears toward the end. These positions should not be interpreted as discrete stages, as the analysis captures only the relative timing of their first appearance and the mechanisms can recur throughout a narrative.

Interestingly, this temporal organization differs from the structural pattern described above. Threat construction is strongly and independently associated with action orientation, yet it does not typically precede action within narratives. In other words, strong structural dependence does not imply temporal mediation. Threat and action can remain tightly coupled even when action-oriented rhetoric appears first. For example, early calls to exclude, ban, punish, or restrict an outgroup are subsequently justified through claims that the group poses a threat.

Table~\ref{tab:guiding-example-temporal} provides an illustrative example of these mechanisms from an anti-trans narrative. Its earliest posts are dominated by boundary construction, including claims that deny transgender people membership in the gender category they claim, followed by explicit negative evaluation. Only later do the narrative’s first calls for punitive action and constructions of transgender people as a threat appear, with dehumanization and scapegoating disproportionately concentrated later still.

% Three statistically distinct temporal tiers emerge consistently across methods. Negative evaluation, boundary construction, and action orientation form an early tier; dehumanization and threat construction form a later tier; and scapegoating occupies a final tier on its own.

% Ideally replace the current single-panel adjusted-rank figure with
% a two-panel raw/adjusted rank figure.

\section{Discussion}
When mechanisms of group-directed hostile rhetoric that are usually studied separately are measured together, a broader organization becomes visible. The six mechanisms are neither independent rhetorical devices nor components of a single fixed pathway. Structurally, most of their predictive dependence is concentrated in a small set of relationships: seven of the fifteen possible edges recover 97.8\% of the predictive gain available to the fully connected model.  %Temporally, the mechanisms show a different but similarly regular organization, separating into early, later, and latest regions as narratives unfold. 
The mechanisms also show a temporal organization, separating into early, later, and latest regions as narratives unfold.

Boundary construction occupies a central role in the rhetoric, connecting broadly across the recovered backbone. This suggests that drawing a salient distinction between an outgroup and the ingroup, construing that outgroup as threatening, and advocating action against it form an especially narrative device of hostile rhetoric. 
Threat construction occupies a similarly prominent position in this system. Its importance is not simply a matter of how often threat language appears. Threat participates in the two strongest conditional relationships we recover, linking most strongly with boundary construction and action orientation. These relationships account for a large share of the predictive structure among all six mechanisms. 
By contrast, relationships centered on dehumanization and moral exclusion account for a smaller share of the independent predictive structure in this corpus.

The temporal results add nuance to the  structural picture. Threat construction is strongly associated with action orientation within posts, yet it does not typically precede action as narratives develop. Action-oriented rhetoric often appears early in narratives, while threat construction tends to appear later. Strong structural dependence  does not imply directionality: indeed, we find that a narrative can begin with calls to exclude, punish, or restrict a target group and only later supply threat-based language that reinforces or justifies those prescriptions.

Existing theories help make sense of different parts of this pattern, but no single theoretical pathway accounts for the organization in full. Relationships emphasized by threat-mediated accounts align especially closely with the recovered structural core: the boundary--threat and threat--action relationships are among the most informative in the system, and the corresponding two-edge structure performs better than any other structure of the same size. At the same time, the temporal sequence implied by that pathway is only partially recovered, because threat does not reliably precede action. 
Moral-exclusion accounts capture other relationships in the system, particularly the connection between group differentiation and dehumanization, but those relationships explain substantially less of the overall structural dependence and their proposed temporal sequence is not recovered as a whole. We therefore interpret these theories less as competing descriptions of the entire system than as accounts of particular relationships that become more or less prominent when their mechanisms are measured together.

For computational social science, this points beyond treating harmful-language categories as separate detection tasks: models that identify dehumanization, threat, blame, or action orientation independently can miss the organization that becomes visible only when those mechanisms are measured together. The sparse structure recovered here suggests that much of this organization can be represented with relatively few relationships, while the temporal results show that the order in which mechanisms enter narratives provides additional information not captured by post-level co-occurrence alone. These patterns could support future work on narrative forecasting, early moderation strategies to prevent hostile rhetoric from growing into mistreatment and incitement, and models that represent combinations of mechanisms rather than isolated labels.

\vspace{3pt}
\noindent
\textbf{Limitations.} 
Our results describe statistical organization in observed rhetoric and should not be interpreted as evidence of causal progression among underlying psychological processes. The mechanisms we measure are linguistic manifestations of broader theoretical constructs, and the same rhetorical expression may reflect a speaker's own beliefs, an attempt to influence an audience, or both. Automated annotation and narrative construction introduce additional measurement choices, although the primary structural findings remain stable under substantial perturbation of the automated labels. Finally, the corpus covers TikTok, Truth Social, and Twitter/X during the 2024 U.S. presidential election; whether the same organization appears in other political contexts, platforms, languages, or forms of group-directed hostility remains an open question.

% \vspace{3pt}
% \noindent
% \textbf{Limitations.} Our results describe observed statistical patterns in harmful rhetoric without inferring direct causal links between mechanisms. The theoretical accounts we test are broader than the specific mechanism relationships used here, so our comparisons should be read as tests of particular structural and temporal claims rather than of entire theoretical traditions. The analysis also depends on automated mechanism annotation and on our construction of narratives, both of which introduce measurement choices that can affect the recovered relationships and timing patterns. Finally, the data come from TikTok, Truth Social, and Twitter/X during the 2024 U.S. election, so the extent to which the same organization appears in other platforms, political contexts, or forms of harmful discourse remains an open question.

\vspace{3pt}
\noindent
\textbf{Future work.} A natural next step is to test whether the same organization appears in other settings. The temporal analysis can also be extended beyond first appearance, and future work should study recurrence, persistence, and transitions among mechanisms as narratives evolve. These patterns may be useful for platform and community interventions as well; systems can use changes in mechanism composition to identify which narratives are moving toward more threatening forms of discourse. Finally, future work that separates the rhetorical manifestation of a mechanism from its actual psychological function would offer a rich new direction for future work and could help distinguish rhetoric from belief formulation.

 % Finally, separating the rhetorical manifestation of a mechanism from its psychological function opens another direction for future work: combining discourse analysis with experiments could help distinguish language that expresses an author's underlying beliefs from language intended to shape the beliefs or behavior of an audience.

% \vspace{3pt}
% \noindent
% \textbf{Future work.} A natural next step is to test whether the structural and temporal organization recovered here holds in other contexts. The temporal analysis can also be extended  to examine recurrence, persistence, and transitions among mechanisms as narratives evolve. The empirical structure recovered here also gives us a basis for testing whether particular combinations of mechanisms or temporal trajectories improve tasks like narrative forecasting and moderation prioritization.

\section{Conclusion}
We studied rhetorical manifestations of social processes creating intergroup hostility. When these rhetorical mechanisms are measured together, a broader organization becomes visible. Most of the structural dependence among the six mechanisms is concentrated in a small and reproducible set of relationships centered on boundary construction and threat construction, and their first appearances follow a distinct and stable temporal organization across narratives.

Group-directed hostile rhetoric is best understood as a system of interdependent mechanisms. Its structure is sparse but highly organized, with most of the predictive dependence concentrated in a small set of relationships, and its temporal development follows a separate and similarly regular ordering across narratives. Existing theories capture important parts of this system, but the larger organization only becomes visible when their mechanisms are examined together.

% \section{Conclusion}

% Harmful rhetoric has often been studied through theories that isolate particular mechanisms---group boundaries, threat, blame, dehumanization, incitement---and treat them as largely separate explanatory objects. Our results show that these mechanisms instead form a highly regular empirical system. Structurally, nearly all of the predictive organization among the six mechanisms is concentrated in a small set of relationships centered on boundary construction and threat construction, and this structure is recovered across models, cross-validation folds, and platforms. Temporally, the mechanisms also occupy stable early, intermediate, and late positions as narratives unfold.

% Existing theories capture this organization very unevenly: accounts emphasizing boundary construction, threat, and harmful action recover much of the system's structural core, while others describe substantially narrower regions; none cleanly reproduces the broader temporal organization observed across narratives. The major theoretical traditions are therefore best understood as partial views of a larger empirical system whose organization becomes visible when their mechanisms are studied together.

\subsection{Acknowledgments}
We thank the annotators for their assistance with annotation across multiple rounds. We also thank Justin Freking for his help with data storage, Noah Eibl for his support, and Kevin Gerard for his helpful feedback.

\bibliography{references}
\clearpage

\newcommand{\answerYes}[1]{\textcolor{blue}{#1}} 
\newcommand{\answerNo}[1]{\textcolor{teal}{#1}} 
\newcommand{\answerNA}[1]{\textcolor{gray}{#1}} 
\newcommand{\answerTODO}[1]{\textcolor{red}{#1}} 
%
% These are recommended to typeset listings but not required. See the subsubsection on listing. Remove this block if you don't have listings in your paper.
\lstset{%
	basicstyle={\footnotesize\ttfamily},% footnotesize acceptable for monospace
	numbers=left,numberstyle=\footnotesize,xleftmargin=2em,% show line numbers, remove this entire line if you don't want the numbers.
	aboveskip=0pt,belowskip=0pt,%
	showstringspaces=false,tabsize=2,breaklines=true}
\floatstyle{ruled}
\newfloat{listing}{tb}{lst}{}
\floatname{listing}{Listing}

\subsection{Paper Checklist to be included in your paper}

\begin{enumerate}

\item For most authors...
\begin{enumerate}
    \item  Would answering this research question advance science without violating social contracts, such as violating privacy norms, perpetuating unfair profiling, exacerbating the socio-economic divide, or implying disrespect to societies or cultures?
    \answerYes{Yes, we study aggregate rhetorical patterns in public social-media discourse. Results are reported at the mechanism and narrative levels.}
  \item Do your main claims in the abstract and introduction accurately reflect the paper's contributions and scope?
    \answerYes{Yes, the abstract and introduction describe the paper as an empirical study of the linguistic manifestations of harmful-rhetoric mechanisms (and note how this is different from claims about underlying psychological states or causal processes).}
    
    \item Do you clarify how the proposed methodological approach is appropriate for the claims made? 
    \answerYes{Yes, we evaluate each major claim from multiple complementary angles and explain why each angle is used. Mechanism annotations are validated against human judgments and across candidate models; structural relationships are evaluated using multiple methods (held-out predictive contribution, alternative network recovery, and platform-level replication); and temporal organization is evaluated through adjusted first-appearance estimates, bootstrap comparisons, and contains robustness checks. We also limit our claims to the statistical organization of observed rhetoric.}
    
\item Do you clarify what are possible artifacts in the data used, given population-specific distributions?
    \answerYes{Yes, we discuss differences in platform composition, mechanism prevalence, automated annotation, narrative construction, and the election-specific sampling frame. We also provide platform-specific and additional robustness analyses.}
    
    \item Did you describe the limitations of your work?
    \answerYes{Yes, the paper has a limitations discussion covering measurement, annotation, narrative construction, causal interpretation, platform coverage, and the study's context.}
    
    \item Did you discuss any potential negative societal impacts of your work?
    \answerYes{Yes, we discuss risks associated with automated classification of hostile rhetoric. This includes overgeneralization, erroneous labeling, and inappropriate use of the framework for individual-level judgments.}
    
    \item Did you discuss any potential misuse of your work?
    \answerYes{Yes, we note that methods for identifying and organizing hostile rhetoric could be repurposed for surveillance or moderation without appropriate review.}
    
    \item Did you describe steps taken to prevent or mitigate potential negative outcomes of the research, such as data and model documentation, data anonymization, responsible release, access control, and the reproducibility of findings?
    \answerYes{Yes, we report aggregate results, we do not release identifying user-level information, and we provide code and methodological details needed to reproduce the analyses.}

    \item Have you read the ethics review guidelines and ensured that your paper conforms to them?
    \answerYes{Yes.}
\end{enumerate}

\item Additionally, if your study involves hypotheses testing...
\begin{enumerate}
  \item Did you clearly state the assumptions underlying all theoretical results?
    \answerYes{Yes, we state how each theory is operationalized as relationships among observable rhetorical mechanisms and we also make clear that these operationalizations concern linguistic manifestations rather than latent psychological processes.}

    \item Have you provided justifications for all theoretical results?
    \answerYes{Yes, the theory-derived relationships and temporal expectations are grounded in the cited social-science literature; empirical conclusions are supported by held-out structural and narrative-level temporal analyses.}
    
    \item Did you discuss competing hypotheses or theories that might challenge or complement your theoretical results?
    \answerYes{Yes, the paper explicitly compares expectations from multiple theoretical traditions and examines where they capture complementary or different portions of the empirical organization.}

    \item Have you considered alternative mechanisms or explanations that might account for the same outcomes observed in your study?
        \answerYes{Yes, we discuss alternative interpretations of the observed linguistic relationships. We also discuss the difference between expressing a psychological state and attempting to induce one in an audience.}
        
    \item Did you address potential biases or limitations in your theoretical framework?
    \answerYes{Yes, we emphasize that the selected mechanisms operationalize particular theoretical traditions and do not contain all possible forms or explanations of hostile rhetoric.}
    
    \item Have you related your theoretical results to the existing literature in social science?
    \answerYes{Yes, the mechanisms and theory-derived expectations are grounded in research in social science and appropriately cited.}

    \item Did you discuss the implications of your theoretical results for policy, practice, or further research in the social science domain?
        \answerYes{Yes, the discussion discusses  implications for studying hostile rhetoric as an interdependent system. We also discuss directions for future work like connecting rhetorical manifestations with psychological processes and experimental evidence.}
    \end{enumerate}

\item Additionally, if you are including theoretical proofs...
\begin{enumerate}
  \item Did you state the full set of assumptions of all theoretical results?
    \answerNA{NA}	
    \item Did you include complete proofs of all theoretical results?
    \answerNA{NA}
    
\end{enumerate}

\item Additionally, if you ran machine learning experiments...
\begin{enumerate}
  \item Did you include the code, data, and instructions needed to reproduce the main experimental results (either in the supplemental material or as a URL)?
    \answerYes{Yes, code and annotation guidelines for reproducing the analyses are included in the accompanying repository.}

    \item Did you specify all the training details (e.g., data splits, hyperparameters, how they were chosen)?
        \answerYes{Yes, we report the evaluated models, inference settings, validation procedure, and relevant model parameters.}
    
    \item Did you report error bars (e.g., with respect to the random seed after running experiments multiple times)?
        \answerYes{Yes when it is applicable; see Results section.}
        
    \item Did you include the total amount of compute and the type of resources used (e.g., type of GPUs, internal cluster, or cloud provider)?
        \answerYes{Yes, our appendix reports the computational environment and GPU resources used.}
        
    \item Do you justify how the proposed evaluation is sufficient and appropriate to the claims made? 
        \answerYes{Yes, automated annotation is evaluated against a human-adjudicated benchmark; also, analyses are evaluated across multiple angles for robustness.}
        
    \item Do you discuss what is ``the cost`` of misclassification and fault (in)tolerance?
        \answerYes{Yes, we treat model predictions as measurements, validate them against human annotations, and we also examine the sensitivity of the primary structural findings to annotation error.}

\end{enumerate}

\item Additionally, if you are using existing assets (e.g., code, data, models) or curating/releasing new assets, \textbf{without compromising anonymity}...
\begin{enumerate}
  \item If your work uses existing assets, did you cite the creators?
    \answerYes{Yes, see Data section.}
    \item Did you mention the license of the assets?
    \answerYes{Yes, we follow the licenses and usage terms of the underlying datasets and open-weight models.}  
    \item Did you include any new assets in the supplemental material or as a URL?
    \answerYes{Yes, we provide code and reproducibility materials through the linked repository.}
    \item Did you discuss whether and how consent was obtained from people whose data you're using/curating?
    \answerNA{NA}
    \item Did you discuss whether the data you are using/curating contains personally identifiable information or offensive content?
    \answerYes{Yes, analyses are conducted on deidentified info and identifying user information is not reported in the paper.}
    
    \item If you are curating or releasing new datasets, did you discuss how you intend to make your datasets FAIR (see \citet{fair})?
    \answerNA{NA}
    
    \item If you are curating or releasing new datasets, did you create a Datasheet for the Dataset (see \citet{gebru2021datasheets})? 
    \answerNA{NA}
    
    \end{enumerate}

\item Additionally, if you used crowdsourcing or conducted research with human subjects, \textbf{without compromising anonymity}...
\begin{enumerate}
  \item Did you include the full text of instructions given to participants and screenshots?
    \answerNA{NA}

    \item Did you describe any potential participant risks, with mentions of Institutional Review Board (IRB) approvals?
    \answerNA{NA}
    
    \item Did you include the estimated hourly wage paid to participants and the total amount spent on participant compensation?
    \answerNA{NA}
    
    \item Did you discuss how data is stored, shared, and deidentified?
    \answerYes{Yes, we do not release identifying user-level information.}

    \end{enumerate}

\end{enumerate}

\clearpage
\appendix
\section{Appendix}

\section{Inter-Annotator Agreement}

Table~\ref{tab:human_agreement} reports agreement between the two annotators on the initial 100-item sample for each mechanism, before adjudication. Disagreements were subsequently reviewed to establish the reference label.

\begin{table}[h]
\centering
\small
\caption{Human inter-annotator agreement prior to adjudication.}
\label{tab:human_agreement}
\begin{tabular}{lrr}
\toprule
Mechanism & Raw agreement & Cohen's $\kappa$ \\
\midrule
Action orientation      & .930 & .497 \\
Boundary construction   & .890 & .747 \\
Dehumanization          & .960 & .823 \\
Negative evaluation     & .930 & .733 \\
Scapegoating            & .850 & .427 \\
Threat construction     & .870 & .628 \\
\midrule
Mean                     & .905 & .643 \\
\bottomrule
\end{tabular}
\end{table}

\section{Validation Set Composition}
\label{app:validation_composition}

The final evaluation set contains exactly 100 post--mechanism pairs for each mechanism. 

\begin{table}[h]
\centering
\small
\caption{Human reference-label composition of the final validation set.}
\label{tab:validation_composition}
\begin{tabular}{lrrr}
\toprule
Mechanism & Positive & Negative & Total \\
\midrule
Action orientation      & 36 & 64 & 100 \\
Boundary construction   & 42 & 58 & 100 \\
Dehumanization          & 33 & 67 & 100 \\
Negative evaluation     & 46 & 54 & 100 \\
Scapegoating            & 61 & 39 & 100 \\
Threat construction     & 44 & 56 & 100 \\
\midrule
Total                    & 262 & 338 & 600 \\
\bottomrule
\end{tabular}
\end{table}

Platform coverage remained similar across mechanisms. Overall, the final benchmark contains 377 Twitter observations (62.8\%), 156 TikTok observations
(26.0\%), and 67 Truth Social observations (11.2\%).

% \begin{table}[h]
% \centering
% \small
% \caption{Platform composition of the final validation set by mechanism.}
% \label{tab:validation_platforms}
% \begin{tabular}{lrrr}
% \toprule
% Mechanism & Twitter & TikTok & Truth Social \\
% \midrule
% Action orientation      & 63 & 26 & 11 \\
% Boundary construction   & 61 & 26 & 13 \\
% Dehumanization          & 62 & 30 & 8 \\
% Negative evaluation     & 66 & 25 & 9 \\
% Scapegoating            & 63 & 26 & 11 \\
% Threat construction     & 62 & 23 & 15 \\
% \midrule
% Total                    & 377 & 156 & 67 \\
% \bottomrule
% \end{tabular}
% \end{table}

\section{LLM Evaluation Details}
\label{app:llm_annotation}

All candidate models received the same mechanism definitions used by the human annotators and classified the same 600 post--mechanism pairs. Human labels and
the GPT-5 labels used during sampling were not included in the model prompts. Inference used deterministic decoding ($T=0$), and models were instructed to return a binary label. We report macro-averaged results because each of the six mechanisms forms a separate binary annotation task. Table~\ref{tab:qwen_by_mechanism} reports the selected model's performance for each mechanism.

All model evaluation and corpus-scale annotation were run on our institutional GPU cluster using a single NVIDIA RTX A6000 GPU with vLLM and deterministic decoding.

% \begin{table*}[t]
% \centering
% \small
% \caption{Qwen3-14B performance against human reference annotations by mechanism.}
% \label{tab:qwen_by_mechanism}
% \begin{tabular}{lrrrrrr}
% \toprule
% Mechanism & Accuracy & Precision & Recall & F$_1$ &
% Cohen's $\kappa$ & Specificity \\
% \midrule
% Action orientation      & .820 & .688 & .917 & .786 & .636 & .766 \\
% Boundary construction   & .760 & .636 & 1.000 & .778 & .543 & .586 \\
% Dehumanization          & .850 & .714 & .909 & .800 & .683 & .821 \\
% Negative evaluation     & .750 & .672 & .891 & .766 & .509 & .630 \\
% Scapegoating            & .790 & .770 & .934 & .844 & .530 & .564 \\
% Threat construction     & .810 & .719 & .932 & .812 & .626 & .714 \\
% \midrule
% Macro average           & .797 & .700 & .931 & .798 & .588 & .680 \\
% \bottomrule
% \end{tabular}
% \end{table*}

\begin{table}[ht]
\centering
\small
\caption{Qwen3-14B performance against human reference annotations by mechanism.}
\label{tab:qwen_by_mechanism}
\begin{tabular}{lrrrr}
\toprule
Mechanism & Precision & Recall & F$_1$ & Cohen's $\kappa$ \\
\midrule
Action orientation      & .688 & .917 & .786 & .636 \\
Boundary construction   & .636 & 1.000 & .778 & .543 \\
Dehumanization          & .714 & .909 & .800 & .683 \\
Negative evaluation     & .672 & .891 & .766 & .509 \\
Scapegoating            & .770 & .934 & .844 & .530 \\
Threat construction     & .719 & .932 & .812 & .626 \\
\midrule
Macro average           & .700 & .931 & .798 & .588 \\
\bottomrule
\end{tabular}
\end{table}

% We also measured agreement among candidate models.

% \begin{table}[h]
% \centering
% \small
% \caption{Agreement between Qwen3-14B and other candidate models.}
% \label{tab:intermodel_agreement}
% \begin{tabular}{lrr}
% \toprule
% Model & Raw agreement & Cohen's $\kappa$ \\
% \midrule
% Gemma-3-12B      & .818 & .615 \\
% Gemma-3-4B       & .742 & .431 \\
% Llama-3.1-8B     & .642 & .168 \\
% Gemma-3-1B       & .580 & .010 \\
% \bottomrule
% \end{tabular}
% \end{table}

\section{Narrative Construction and Cross-Platform Coverage}
\label{app:narratives}

Following prior work, we define a narrative as a collection of posts centered on the same underlying issue, event, or claim, forming a coherent unit of discourse that can evolve across time and platforms~\citep{hanley2024partial, gerard2026cross,gerard2026bridging}. Narrative clusters are constructed from semantically normalized claims so that equivalent expressions can be grouped despite differences in wording or platform-specific style.

The temporal analysis in this paper uses 2,513 narratives with sufficient mechanism activity for first-appearance estimation. Of these, 2,492 (99.2\%) span at least two of the three platforms in our corpus.

As an example, one narrative tracks the attempted assassination of Donald Trump involving Ryan Routh. On TikTok, posts include claims such as ``Ryan Routh may be part of a bigger conspiracy.'' On Twitter/X, the same narrative appears as ``Ryan Routh has been charged with attempted assassination of Donald Trump,'' while Truth Social contains claims such as ``Ryan Wesley Routh was the gunman in a Trump assassination attempt.'' Although the wording and framing differ across platforms, the posts refer to the same underlying event and narrative.

The narrative construction itself has been validated independently in prior work~\citep{gerard2026cross}. Human annotators evaluating whether claims grouped into the same cluster referred to the same narrative achieved 91.5\% accuracy. Second, the claim-normalization step was validated at 91.5\% semantic-preservation accuracy. Finally, claim normalization reduces source-platform prediction accuracy to near-chance, suggesting that the claims extracted from each post before narrative clustering retain little platform-specific linguistic signal~\cite{gerard2026cross}, and thus the resulting narratives collected indeed are unbiased to certain platform semantic structure.

\section{Platform Heterogeneity in Mechanism Prevalence}
\label{app:platform-heterogeneity}

Mechanism prevalence varies across TikTok, Twitter/X, and Truth Social (Table~\ref{tab:platform-heterogeneity}). All six omnibus tests are statistically significant due to the size of the corpus; however, the corresponding effect sizes are small, with Cramér's $V$ ranging from .006 to .026 (well below conventional thresholds for even a small association). No platform is uniformly higher across mechanisms: for example, Truth Social has the highest prevalence of threat construction, whereas TikTok has the highest prevalence of scapegoating, negative evaluation, and dehumanization.

% \begin{table*}[ht]
% \centering
% \caption{Prevalence of harmful-rhetoric mechanisms by platform. Values indicate the percentage of posts containing each mechanism. Cramér's $V$ summarizes the magnitude of platform heterogeneity; $q$ values are Benjamini--Hochberg adjusted across mechanisms.}
% \label{tab:platform-heterogeneity}
% \begin{tabular}{lrrrrr}
% \toprule
% \textbf{Mechanism} &
% \textbf{TikTok} &
% \textbf{Twitter/X} &
% \textbf{Truth Social} &
% \textbf{$V$} &
% \textbf{$q$} \\
% \midrule
% Boundary construction & 2.23\% & 3.03\% & 2.12\% & 0.025 & $<.001$ \\
% Threat construction & 1.04\% & 1.49\% & 1.51\% & 0.018 & $<.001$ \\
% Scapegoating & 0.24\% & 0.18\% & 0.17\% & 0.006 & $<.001$ \\
% Negative evaluation & 0.65\% & 0.61\% & 0.35\% & 0.012 & $<.001$ \\
% Dehumanization & 0.57\% & 0.47\% & 0.27\% & 0.013 & $<.001$ \\
% Action orientation & 1.57\% & 1.63\% & 1.34\% & 0.007 & $<.001$ \\
% \bottomrule
% \end{tabular}
% \end{table*}
\begin{table}[ht]
\centering
\small
\caption{Prevalence of harmful-rhetoric mechanisms by platform. Values indicate the percentage of posts containing each mechanism. Cramér's $V$ summarizes the magnitude of platform heterogeneity.}
\label{tab:platform-heterogeneity}

\resizebox{\columnwidth}{!}{%
\begin{tabular}{lrrrr}
\toprule
\textbf{Mechanism} &
\textbf{TikTok} &
\textbf{Twitter/X} &
\textbf{Truth Social} &
\textbf{$V$} \\
\midrule
Boundary construction & 2.23\% & 3.03\% & 2.12\% & 0.025 \\
Threat construction   & 1.04\% & 1.49\% & 1.51\% & 0.018 \\
Scapegoating          & 0.24\% & 0.18\% & 0.17\% & 0.006 \\
Negative evaluation   & 0.65\% & 0.61\% & 0.35\% & 0.012 \\
Dehumanization        & 0.57\% & 0.47\% & 0.27\% & 0.013 \\
Action orientation    & 1.57\% & 1.63\% & 1.34\% & 0.007 \\
\bottomrule
\end{tabular}%
}

\end{table}

\begin{table}[h!]
\centering
\small
\caption{Adjusted associations between harmful-rhetoric mechanisms and engagement. Outcomes are within-platform standardized log engagement measures. Models include all six mechanisms simultaneously, platform and month fixed effects, and a leave-one-out measure of each author's typical engagement on that platform.}
\label{tab:engagement-robustness}
\begin{tabular}{lrrrr}
\toprule
Mechanism & Likes & Replies & Reposts & Views \\
\midrule
Boundary construction &  0.001 &  0.003 &  0.004 & -0.018 \\
Threat construction   & -0.008 &  0.001 &  0.020 & -0.035 \\
Scapegoating          &  0.037 &  0.026 &  0.067 &  0.011 \\
Negative evaluation   & -0.009 & -0.009 &  0.000 & -0.018 \\
Dehumanization        & -0.023 & -0.041 & -0.015 & -0.030 \\
Action orientation    &  0.006 &  0.013 &  0.036 & -0.021 \\
\bottomrule
\end{tabular}
\end{table}

\begin{table*}[th]
\caption{Temporal guiding example: mechanism first-appearance positions within a single representative narrative.}
\centering
\small
\renewcommand{\arraystretch}{1.08}
\setlength{\tabcolsep}{4pt}

\begin{tabular}{@{}llp{4.7cm}p{6.7cm}@{}}
\toprule
\textbf{Position} & \textbf{Mechanism} & \textbf{Representative claim} & \textbf{Interpretation} \\
\midrule

0.004 &
Boundary construction &
``Cisgender individuals who identify as women are actually men.'' &
Establishes a categorical boundary almost immediately by denying transgender people membership in the gender category they claim. \\[4pt]

0.077 &
Negative evaluation &
``Transgender individuals are a `weak bad seed.''' &
Moves from categorical distinction to explicit derogatory evaluation of the target. \\[4pt]

0.310 &
Action orientation &
``Liberals and trans people should be imprisoned and possibly be deported.'' &
Introduces an explicit punitive prescription directed at transgender people. \\[4pt]

0.326 &
Threat construction &
``Mentally ill trans folks are a threat to people's lives.'' &
Explicitly constructs transgender people themselves as dangerous to others. \\[4pt]

0.437 &
Dehumanization &
``[An individual described as] a transgender illegal alien eats cats and dogs.'' &
Introduces a more degrading portrayal of the target, alongside other harmful mechanisms. \\[4pt]

0.448 &
Scapegoating &
``The government is paying for transgender illegal aliens to have sex changes in prison.'' &
Incorporates the target into a broader political grievance concerning government resources, immigration, and perceived preferential treatment. \\

\bottomrule
\end{tabular}

\label{tab:guiding-example-temporal}
\end{table*}

\begin{figure}[t]
    \centering

\pgfplotstableread{
x y meta
0 0 0.0000
1 0 4.3083
2 0 1.3624
3 0 2.7579
4 0 3.2224
5 0 3.9161

0 1 4.3083
1 1 0.0000
2 1 3.9026
3 1 0.3412
4 1 2.3167
5 1 -1.0313

0 2 1.3723
1 2 3.9088
2 2 0.0000
3 2 2.0938
4 2 1.2294
5 2 0.2808

0 3 2.8035
1 3 0.2471
2 3 2.1677
3 3 0.0000
4 3 -0.1363
5 3 0.6255

0 4 3.1997
1 4 2.4038
2 4 1.0403
3 4 0.0139
4 4 0.0000
5 4 0.6790

0 5 3.9225
1 5 -1.0838
2 5 0.4053
3 5 0.6353
4 5 0.7396
5 5 0.0000
}\heatmapdata

\resizebox{\linewidth}{!}{%
\begin{tikzpicture}
\begin{axis}[
    width=11cm,
    height=9cm,
    colormap={PeachWhiteBlue}{
        rgb255(0cm)=(255,172,129)
        rgb255(3cm)=(255,255,255)
        rgb255(6cm)=(0,117,242)
    },
    colorbar,
    point meta min=-6,
    point meta max=6,
    colorbar style={
        ytick={-6,-3,0,3,6},
        yticklabels={0.016,0.125,1,8,64},
        ylabel={Odds ratio},
    },
    y dir=reverse,
    xtick={0,...,5},
    ytick={0,...,5},
    xticklabels={
        Boundary,
        Threat,
        Action orientation,
        Negative evaluation,
        Dehumanization,
        Scapegoating
    },
    yticklabels={
        Boundary,
        Threat,
        Action orientation,
        Negative evaluation,
        Dehumanization,
        Scapegoating
    },
    x tick label style={
        rotate=40,
        anchor=east,
        font=\small
    },
    y tick label style={font=\small},
    xlabel={Target mechanism (predicted)},
    ylabel={Source mechanism (predictor)},
    axis on top,
    enlargelimits=false,
    xmin=-0.5,
    xmax=5.5,
    ymin=-0.5,
    ymax=5.5,
    tick align=outside,
    axis line style={draw=none},
    major tick length=2pt,
]

\addplot[
    matrix plot*,
    mesh/rows=6,
    mesh/cols=6,
    point meta=explicit
]
table[meta index=2] \heatmapdata;

\node[font=\bfseries\small]
    at (axis cs:0,0) {--};

\node[font=\bfseries\small,white]
    at (axis cs:1,0) {19.81};

\node[font=\bfseries\small,black]
    at (axis cs:2,0) {2.57};

\node[font=\bfseries\small,black]
    at (axis cs:3,0) {6.76};

\node[font=\bfseries\small,white]
    at (axis cs:4,0) {9.33};

\node[font=\bfseries\small,white]
    at (axis cs:5,0) {15.10};

\node[font=\bfseries\small,white]
    at (axis cs:0,1) {19.81};

\node[font=\bfseries\small]
    at (axis cs:1,1) {--};

\node[font=\bfseries\small,white]
    at (axis cs:2,1) {14.96};

\node[font=\bfseries\small,black]
    at (axis cs:3,1) {1.27};

\node[font=\bfseries\small,black]
    at (axis cs:4,1) {4.98};

\node[font=\bfseries\small,black]
    at (axis cs:5,1) {0.49};

\node[font=\bfseries\small,black]
    at (axis cs:0,2) {2.59};

\node[font=\bfseries\small,white]
    at (axis cs:1,2) {15.02};

\node[font=\bfseries\small]
    at (axis cs:2,2) {--};

\node[font=\bfseries\small,black]
    at (axis cs:3,2) {4.27};

\node[font=\bfseries\small,black]
    at (axis cs:4,2) {2.35};

\node[font=\bfseries\small,black]
    at (axis cs:5,2) {1.21};

\node[font=\bfseries\small,black]
    at (axis cs:0,3) {6.98};

\node[font=\bfseries\small,black]
    at (axis cs:1,3) {1.19};

\node[font=\bfseries\small,black]
    at (axis cs:2,3) {4.49};

\node[font=\bfseries\small]
    at (axis cs:3,3) {--};

\node[font=\bfseries\small,black]
    at (axis cs:4,3) {0.91};

\node[font=\bfseries\small,black]
    at (axis cs:5,3) {1.54};

\node[font=\bfseries\small,white]
    at (axis cs:0,4) {9.19};

\node[font=\bfseries\small,black]
    at (axis cs:1,4) {5.29};

\node[font=\bfseries\small,black]
    at (axis cs:2,4) {2.06};

\node[font=\bfseries\small,black]
    at (axis cs:3,4) {1.01};

\node[font=\bfseries\small]
    at (axis cs:4,4) {--};

\node[font=\bfseries\small,black]
    at (axis cs:5,4) {1.60};

\node[font=\bfseries\small,white]
    at (axis cs:0,5) {15.16};

\node[font=\bfseries\small,black]
    at (axis cs:1,5) {0.47};

\node[font=\bfseries\small,black]
    at (axis cs:2,5) {1.32};

\node[font=\bfseries\small,black]
    at (axis cs:3,5) {1.55};

\node[font=\bfseries\small,black]
    at (axis cs:4,5) {1.67};

\node[font=\bfseries\small]
    at (axis cs:5,5) {--};

\end{axis}
\end{tikzpicture}%
}

\caption{
Conditional associations among the six mechanisms of hostile rhetoric. Each cell reports the odds ratio for predicting the target mechanism (column) from the source mechanism (row) while conditioning on the remaining four mechanisms. Color represents the log$_2$ odds ratio.
}
\label{fig:conditional-odds}

\end{figure}
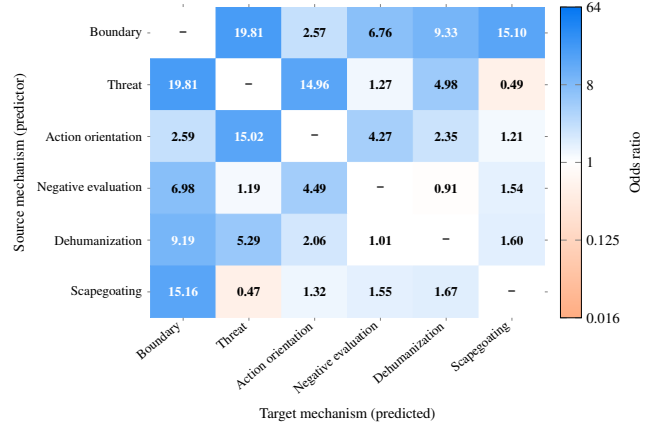

% \begin{figure}[t]
%     \centering
%     \includegraphics[width=\columnwidth]{AnonymousSubmission/LaTeX/figs/backbone_graph_v2.pdf}
%     \caption{
%     Empirical backbone of harmful-rhetoric mechanisms. Node positions were initialized from a spring layout weighted by predictive weight share and adjusted only to remove label overlap; distances do not encode relationship strength or similarity.
%     }
%     \label{fig:backbone-structure}
% \end{figure}

% \clearpage

\section{Structural Estimation Details}
\label{app:structural_methods}

The main analysis examines relationships among six binary rhetorical mechanisms. Since different measures of dependence can emphasize different properties of the joint distribution, we evaluate the system using three complementary methods: held-out predictive dependence, conditional logistic associations, and Bayesian-network structure learning.

\vspace{3pt}
\noindent
\textbf{Held-Out Predictive Scoring}

Let $Y_j$ denote one of the six binary mechanisms and let $\mathcal{N}_j(G)$ denote the mechanisms connected to $Y_j$ under candidate graph $G$. For each target mechanism, we fit a logistic regression using its graph neighbors as predictors,
\[
P(Y_j=1 \mid \mathcal{N}_j(G))
=
\sigma\left(
\beta_{0j}
+
\sum_{k \in \mathcal{N}_j(G)}
\beta_{jk}Y_k
\right),
\]
where $\sigma(\cdot)$ is the logistic function. Models are fit on training posts and evaluated on held-out posts using conditional log-likelihood. A graph's score is the average held-out conditional log-likelihood across the six target mechanisms.
We normalize predictive performance between two references: an independent structure containing no edges and an all-edge structure containing all 15 possible pairwise relationships. For candidate structure $G$, the fraction of available predictive gain recovered is

\[
R(G)
=
\frac{
\mathrm{CLL}(G)-\mathrm{CLL}(G_{\mathrm{ind}})
}{
\mathrm{CLL}(G_{\mathrm{sat}})-\mathrm{CLL}(G_{\mathrm{ind}})
}.
\]

$R(G)=0$ corresponds to the independent model and $R(G)=1$ to the fully connected model.

Since the theoretical structures are undirected, an edge between two mechanisms allows each mechanism to enter the other's predictive model. The resulting score therefore measures symmetric structural dependence.

\vspace{3pt}
\noindent
\textbf{Controlling for Structural Complexity}
Predictive fit generally increases as relationships are added. We therefore benchmark each theory-derived structure against every alternative graph with the same number of edges. We score each candidate using the same held-out procedure and report the theory's rank, percentile, and fraction of the best achievable predictive gain at that edge budget.

\vspace{3pt}
\noindent
\textbf{Recovering the Empirical Predictive Backbone}

We next recover relationships directly from the data . For each ordered pair of mechanisms $Y_k \rightarrow Y_j$, we compare two predictive models for target $Y_j$. The full model contains all five other mechanisms, whereas the reduced model omits $Y_k$. We define the unique predictive contribution of $Y_k$ to $Y_j$ as

\[
\Delta_{k\rightarrow j}
=
\mathcal{L}_{-k}
-
\mathcal{L}_{\mathrm{full}},
\]

where $\mathcal{L}$ denotes held-out log loss. Positive values therefore indicate that removing $Y_k$ worsens held-out prediction of $Y_j$ after the remaining mechanisms have already been included.

Since the structural analysis is undirected, we combine the two directional contributions for each pair:

\[
w_{jk}
=
\max\left(
0,
\frac{
\Delta_{j\rightarrow k}
+
\Delta_{k\rightarrow j}
}{2}
\right).
\]

Edge weights are estimated using five-fold stratified cross-validation.

\vspace{3pt}
\noindent
\textbf{Conditional Association}

The predictive backbone measures whether one mechanism contributes information about another. To directly describe the magnitude of association, we then estimate a complementary conditional-association model.

For each mechanism $Y_j$, we fit a binomial generalized linear model containing all five remaining mechanisms simultaneously:

\[
\operatorname{logit}P(Y_j=1)
=
\alpha_j
+
\sum_{k\neq j}\beta_{jk}Y_k.
\]

Exponentiating $\beta_{jk}$ gives the conditional odds ratio associated with mechanism $k$ while holding the other four mechanisms fixed. We report these odds ratios and their confidence intervals in Figure~\ref{fig:conditional-odds}.

\vspace{3pt}
\noindent
\textbf{Independent Bayesian-Network Recovery}

We also learn Bayesian networks directly from the joint distribution of the six binary mechanisms and use this as a robustness check. We use hill-climbing structure search with the Bayesian Information Criterion (BIC) as the scoring function. 

Since observational Bayesian-network directions should not be interpreted causally here, we discard edge direction and compare only the learned skeleton with the predictive backbone.

\section{Boundary and Threat are Distinct Mechanisms}

Because boundary construction and threat construction form the strongest relationship in the recovered structure, we examine whether this result could reflect overlap in the annotation definitions rather than a substantive relationship between distinct mechanisms. The two labels are strongly associated, but they are far from interchangeable. Among posts containing either boundary construction or threat construction, 85.0\% contain only one of the two mechanisms; only 15.0\% contain both. Threat construction appears in 19.7\% of boundary-construction posts, while boundary construction appears in 38.8\% of threat-construction posts. Their Jaccard similarity is 0.15 and their $\phi$ coefficient is 0.26. The distinction is also visible qualitatively: boundary-only posts commonly establish or characterize an out-group without describing an immediate danger, whereas threat-only posts describe danger, harm, or risk without constructing a broader in-group/out-group distinction.

\subsection{Engagement and Visibility Robustness}
\label{app:engagement-robustness}

We examine whether differences in platform visibility and engagement could plausibly account for the mechanisms identified in our analysis. Across all social media platforms, every mechanism appears in all ten within-platform engagement deciles, indicating that the mechanisms are not confined to unusually visible content. Since algorithmic ranking and platform-specific exposure can also produce systematic differences in which posts receive attention, we also test whether mechanism presence is associated with engagement after accounting for platform, time period, author-level engagement baselines, and co-occurring mechanisms. As shown in Table~\ref{tab:engagement-robustness}, these adjusted associations are uniformly small: the mean absolute standardized coefficient across all 24 mechanism--engagement combinations is only $0.019$, and the maximum absolute standardized coefficient is $0.067$.

\definecolor{Boundary}{HTML}{0075F2}
\definecolor{Action}{HTML}{4CB5AE}
\definecolor{Threat}{HTML}{ace1af}
\definecolor{Negative}{HTML}{ffa9a3}
\definecolor{Dehuman}{HTML}{b8b8ff}
\definecolor{Scapegoat}{HTML}{EF5B5B}

\definecolor{early_tier}{HTML}{0075F2}
\definecolor{later_tier}{HTML}{4CB5AE}
\definecolor{latest_tier}{HTML}{FFAC81}

\section{Robustness to Left Censoring}
\label{app:left-censoring}
A potential concern with our temporal ordering is that some narratives may already have been developing before they enter our observation window. If so, the observed timing of mechanisms could partly reflect left censoring rather than their relative position within the narrative itself.

We test this by progressively restricting the analysis to narratives whose first observed appearance occurs only after an increasingly long period of prior corpus observation. For each threshold from 0 to 30 days, we repeat the adjusted temporal analysis (specified in the main text).

We find that indeed the temporal organization is highly robust. Relative to the baseline adjusted ordering, the exact six-mechanism ranking remains very similar across all thresholds (mean Spearman $\rho=.95$; minimum $\rho=0.89$). The small changes that do occur are confined to reorderings of mechanisms within the same temporal tier and direct tests of these reorderings show that none are statistically distinguishable after Holm correction. Finally, the broader three-tier organization holds at every  threshold.

\section{Robustness to Annotation Error}
\label{app:annotation_error}

We also test whether the recovered structural organization is sensitive to errors in the automated mechanism labels. To do this, we use a prevalence-preserving perturbation test: for each mechanism, we randomly replace a fraction of positive assignments with an equal number of negative assignments. This preserves the marginal prevalence of each mechanism while disrupting which posts receive the label. We first evaluate replacement rates of 10\%, 25\%, and 50\%; then we use a benchmark-informed condition based on the mechanism-specific precision observed for Qwen3-14B (Table~\ref{tab:qwen_by_mechanism}). For each condition, we repeat the perturbation 100 times and rerun the same structural recovery procedure used in the main analysis.

The recovered organization remains highly stable across perturbation levels (Table~\ref{tab:annotation_error_sensitivity}). The ranking of pairwise predictive relationships is strongly correlated with the unperturbed structure (mean Spearman's $\rho=.883$), and remains so even under the deliberately severe 50\% replacement condition ($\rho=.853$).

\begin{table}[ht]
\centering
\caption{Stability of pairwise structural relationships under prevalence-preserving label perturbation. Values report the mean Spearman correlation between perturbed and unperturbed pairwise predictive edge weights across 100 simulations.}
\label{tab:annotation_error_sensitivity}
\begin{tabular}{lc}
\toprule
Perturbation condition & Mean Spearman's $\rho$ \\
\midrule
10\% replacement & .931 \\
25\% replacement & .903 \\
Qwen-informed & .883 \\
50\% replacement & .853 \\
\bottomrule
\end{tabular}
\end{table}

The two dominant relationships in the main analysis (boundary construction with threat construction and threat construction with action orientation) are especially robust; the instability that does occur is concentrated among the weakest edges near the 95\% backbone-selection threshold.

\begin{figure}[htbp]
\centering

\begin{tikzpicture}

\begin{axis}[
    width=0.98\linewidth,
    height=0.62\linewidth,
    xlabel={Required prior observation (days)},
    ylabel={Adjusted first appearance (\% of narrative)},
    xmin=-1.25,
    xmax=32.5,
    ymin=26,
    ymax=37,
    xtick={0,5,10,15,20,25,30},
    ytick={26,28,30,32,34,36},
    yticklabel={\pgfmathprintnumber{\tick}\%},
    ymajorgrids=true,
    grid style={dashed, gray!30},
    tick align=outside,
    legend style={
        at={(0.5,-0.30)},
        anchor=north,
        draw=none,
        fill=none,
        font=\small,
        cells={align=left},
        legend columns=2
    },
    legend cell align=left,
    clip=false
]

\addplot+[
    solid,
    color=Boundary,
    ultra thick,
    mark=none
]
table[
    x=window_days,
    y expr=100*\thisrow{boundary_construction},
    col sep=comma
] {csvs/adjusted_cold_start_temporal_plot.csv};
\addlegendentry{Boundary construction}

\addplot+[
    solid,
    color=Action,
    ultra thick,
    mark=none
]
table[
    x=window_days,
    y expr=100*\thisrow{action_orientation},
    col sep=comma
] {csvs/adjusted_cold_start_temporal_plot.csv};
\addlegendentry{Action orientation}

\addplot+[
    solid,
    color=Threat,
    ultra thick,
    mark=none
]
table[
    x=window_days,
    y expr=100*\thisrow{threat_construction},
    col sep=comma
] {csvs/adjusted_cold_start_temporal_plot.csv};
\addlegendentry{Threat construction}

\addplot+[
    solid,
    color=Negative,
    ultra thick,
    mark=none
]
table[
    x=window_days,
    y expr=100*\thisrow{negative_evaluation},
    col sep=comma
] {csvs/adjusted_cold_start_temporal_plot.csv};
\addlegendentry{Negative evaluation}

\addplot+[
    solid,
    color=Dehuman,
    ultra thick,
    mark=none
]
table[
    x=window_days,
    y expr=100*\thisrow{dehumanization},
    col sep=comma
] {csvs/adjusted_cold_start_temporal_plot.csv};
\addlegendentry{Dehumanization}

\addplot+[
    solid,
    color=Scapegoat,
    ultra thick,
    mark=none
]
table[
    x=window_days,
    y expr=100*\thisrow{scapegoating},
    col sep=comma
] {csvs/adjusted_cold_start_temporal_plot.csv};
\addlegendentry{Scapegoating}

% Early tier
\draw[
    Boundary,
    thick,
    color=early_tier
]
(axis cs:31.2,30.3)
--
(axis cs:31.8,30.3)
--
(axis cs:31.8,31.8)
--
(axis cs:31.2,31.8);

\node[
    anchor=west,
    text=Boundary,
    color=early_tier,
    font=\small
]
at (axis cs:32.5,31.05)
{Early tier};

% Later tier
\draw[
    Action,
    thick,
    color=later_tier
]
(axis cs:31.2,32.5)
--
(axis cs:31.8,32.5)
--
(axis cs:31.8,33.7)
--
(axis cs:31.2,33.7);

\node[
    anchor=west,
    text=Action,
    color=later_tier,
    font=\small
]
at (axis cs:32.5,33.1)
{Later tier};

% Latest tier
\draw[
    Scapegoat,
    thick,
    color=latest_tier
]
(axis cs:31.2,35.1)
--
(axis cs:31.8,35.1)
--
(axis cs:31.8,36.1)
--
(axis cs:31.2,36.1);

\node[
    anchor=west,
    text=Scapegoat,
    color=latest_tier,
    font=\small
]
at (axis cs:32.5,35.6)
{Latest tier};
% \node[
%     anchor=north west,
%     font=\scriptsize,
%     align=left,
%     fill=white,
%     fill opacity=0.92,
%     text opacity=1,
%     inner sep=3pt
% ]
% at (axis cs:0.5,36.7)
% {
% Max.\ relative shift: 0.81 pp
% };

\end{axis}

\end{tikzpicture}

\caption{
Robustness of adjusted mechanism first-appearance timing to potential left censoring. We progressively require between 0 and 30 days of prior corpus observation before a narrative's first observed appearance. The \textbf{apparent reorderings within the early and later tiers are not statistically distinguishable} after Holm correction, and the three-tier temporal organization remains intact across all 31 thresholds. 
}

\label{fig:left-censoring-robustness}

\end{figure}
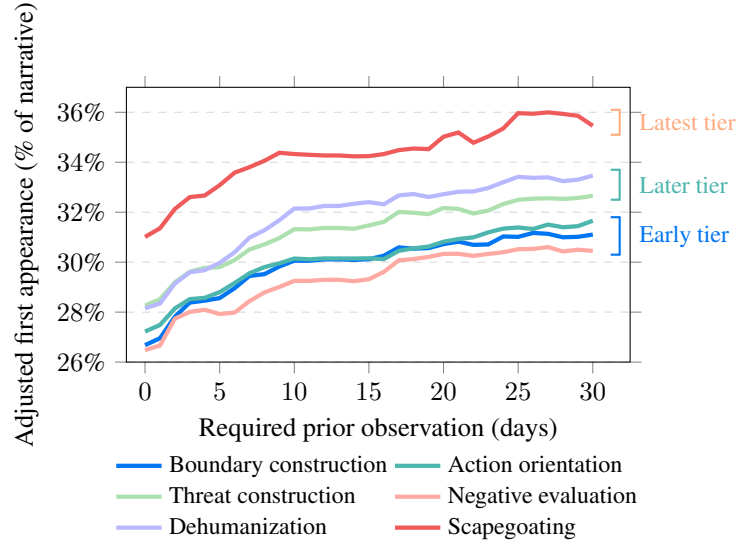

\end{document}